\documentclass[aps,preprint,showkeys]{revtex4-1}
\usepackage{color}
\usepackage{graphicx}
\usepackage{subcaption}
\usepackage{tabularx}
\usepackage[T1]{fontenc}

\begin{document}

\title{Customized determination of stop words using Random Matrix Theory approach}
\author{Bogdan {\L}obodzi\'{n}ski}
\email[]{lobodzinskibogdan@gmail.com}

\date{\today}

\begin{abstract}
The distances between words calculated in word units are studied and compared with the distributions of the 
Random Matrix Theory (RMT). 
It is found that the distribution of distance between the same words can be well described by the single-parameter Brody distribution. 
Using the Brody distribution fit, we found that the distance between given words in a set of texts can show mixed dynamics, coexisting regular and chaotic regimes.  
It is found that distributions correctly fitted by the Brody distribution with a certain goodness of the fit threshold can be identifid as stop words, usually considered as the uninformative part of the text. 
By applying various threshold values for the goodness of fit, we can extract uninformative words from the texts under analysis to the desired extent. 
On this basis we formulate a fully agnostic recipe that can be used in the creation of a customized set of stop words for texts in any language based on words.
\end{abstract}

\keywords{stop words; text mining; NLP processing; Random MAtrix Theory; statistics; information retrieval; word frequency}

\maketitle

\section{Introduction}

Words in natural languages can be used in different ways depending on the purpose of the text.  The grammar, the type of words used, the order of used words and the frequency of their use are different for technical descriptions, personal letters, texts used in social media or in newspaper articles. 
Different types of strategies can be used to emphasize the main purpose of the message, which can be, for example, expression of sentiment, conveying information, opinions or influencing the target audience.   

If the purpose of analysing a text is to extract the information contained in the document, then an important pre-processing step is to locate and remove words that do not affect information. Their elimination directly influences the ultimate result of retrieval and information mining. This kind of noniformative words is commonly understood as stop words in the Natural language processing (NLP) jargon.

The opposite pole is sentiment analysis, where much more important than the information is the relationship between  
words and the main purpose of the text is conveyed by the ordering of words. In this case, words belonging to the generally understood class of stop words may turn out to be informative words. 

In this work, we will focus on the analysis of texts from the point of view of extracting information and, more specifically, on the statistical analysis of the distance between the words in the set of texts and how to use these results to differentiate between important and unimportant words. 
The aim of this analysis is to find a method that will help to identify certain words in a collection of analysed texts that are not informative in terms of meaning, but which by their presence mark the stylistic value of the sentences used 
or constitute so-called special words (times, file names, dates, prefixes, abbreviations, units etc). 
Such a set of words will be defined as stop words in the further part of the work.    

As a source of data for analysis we use Wikipedia articles in English, German and Polish, which are available freely and online \cite{Wikiarticles}.  
For each language we have chosen 20000 documents.

\section{Statistical analysis of distances between words} 

In this work we first examine the possibility that the distribution of average distances (calculated per document) in word units calculated for words from a corpus built from an assumed set of  documents for a given language shows universality of the type predicted by the Random Matrix Theory (RMT). The RMT is widely used in many physical \cite{PhysicalRMT}, biological \cite{BiologyRMY}, financial \cite{FinanceRMT} and other systems. 
To the author's knowledge, NLP is one of the very few domains where this theory has not yet been applied.

The methodology for the preparation of data for analysis is as follows:
\begin{enumerate}
\item for a given language, we create a set of used words, sorting them down by the number of occurrences in all articles.
\item For the first 10000 most frequent words, we calculate the distance between given word in each sentence of the article (without taking into account punctuation and special characters) in units of words. Then, for each article, we calculate the average distance between the words. 
\end{enumerate}

Each word $w_{p}$ is therefore represented by a series of tuples 
\begin{equation}\label{SeriesInitial}
T_{p} = \left\{..., \left(x_{i}, l_{i}\right), \left(x_{i+1}, l_{i+1}\right), ...\right\}
\end{equation}
where 
$x_{k}$ is the average distance (in word s unit) of the word $w_{p}$ in the article ${k}$ and $l_{k}$ stands for the length (in units of words) of the article ${k}$. 
The avearge distances $x_{k}$ from the series $T_{p}$ are standarized with article s lengths ($l_{k}$) as weights ($\hat{T}_{p}$), thus:

\begin{equation}\label{SeriesStandarized}
T_{p} \rightarrow \hat{T}_{p} = \left\{..., \hat{x}_{i}, \hat{x}_{i+1}, ...\right\}
\end{equation}
where $\hat{x}_{k}$ is the standarized average distance of the word $w_{p}$ in the article ${k}$. 

In the next step, the standard procedure of unfolding of the time series is provided ($\hat{T}_{p}$), so that the mean average distance is unity. 
It is assumed that the cumulated, the staircase function $N\left(X\right)$ (event density) can be separated into a smoothed ($N_{smooth}$) and a
fluctuating ($N_{fluct}$) parts 

\begin{equation}\label{Staircase}
N\left( X \right) = \frac{1}{n}\sum_{k} \theta\left(X - \hat{x}_{i}\right) = N_{smooth} ( \hat{x} ) + N_{fluct} ( \hat{x} )  
\end{equation}

The unfolded values $X_{i}$ are obtained by the mapping of the smooth part onto the $\hat{x}_{i}$ using the transformation:
\begin{equation}\label{StaircaseTransformation}
X_{i} \rightarrow X_{i,smooth} \left( \hat{x}_{i} \right) .
\end{equation}
Later on, all unfolded values are detrended \cite{Unfolding} with help of Empirical Mode Decomposition \cite{EMD}.
For the purposes of this analysis we modified part of a code provided by the Python package {\it empyricalRMT} \cite{empyricalRMT}.  

The data prepared in this way can be used to calculate statistical measures, which are the nearest-neighbor distance distribution
$P\left(X\right)dX$ which provides the probability to find $X_{i}$ in the interval  $\left\{ X, X+dX \right\}$. $P\left(X\right)$ is the short range correlations between distances $X$. 

Trying to understand the problem of the distribution of words in the text intuitively, 
one can assume that the distribution of the average distance between a given word counted in units of words in a given document will show two types of behaviour: ordered and irregular. 
This is based on the fact that the analyzed collection of texts was created by different authors and can be attributed to different characteristics (writing styles) expressed mainly in the choice of words and length of sentences .
\begin{itemize}
\item {\it Regular behavior}: 
when the use of the word $A$ does not imply the use of the word $B$, i.e. words $A$ and $B$ can be treated as uncorrelated variables.
This behavior can be expected if a word is used strictly according to grammatical rules and in short sentences.  
\item {\it Irregular behaviour}: 
when the use of word $A$ implies the use of word $B$ (in terms of meaning and position in the sentence or text). Then, words $A$ and $B$ are related to each other and are correlated variables. This can occur with longer words, longer and more complex sentences and documents.
\end{itemize}
The existence of correlations between words will affect the statistical behavior of the average distance between words (spacing) calculated in the analyzed texts.
In practice, we should observe different regimes: regular (integrable) and intermediate between regular and irregular (chaotic). 
Here, we will consider the measured average distances between words as observables of some physical system whose type of behavior we want to determine.    
In their paper Berry and Tabor \cite{Berry} 
formulated and proved a theorem that if, in the classical limit, the dynamics of a quantum system is completely integrable, 
the distribution describing the distance $X$ (spacing) between successive unfolded energy eigenvalues of such a system corresponds to the Poisson distribution 
$P\left(X\right) = exp\left(-X\right)$. That means, we are working with a sequence of uncorrelated random values. 

If a disorder occurs in the system that leads to breaking of the integrability, i.e. word $A$ influences the distribution of word $B$, then we are dealing with behavior leading to a non-Poissonian distribution of the spacing $X$. In the chaotic limit, the distribution of $X$ takes the shape of the Wigner distribution 
$P\left(X\right) = \frac{\pi}{2} X e^{\left(-\frac{\pi}{4}X^{2}\right)}$ \cite{WignerSurmise}.

The semiclassical distribution describing the statistics of spacing $X$ in the case of co-existence of regular and irregular behaviors in dynamical systems is well described by the one-parameter Brody distribution \cite{Brody}.

\begin{equation}\label{BrodyFormula}
P_{q}\left(X\right) = \left(1 + q\right) B\left(q\right) X^{q} exp\left( - B\left(q\right) X^{\left(1+q\right)}\right)
\end{equation}
with 
$B\left(q\right) = \left\{\Gamma\left[\frac{\left(2+q\right)}{\left(1+q\right)}\right]\right\}^{\left(1+q\right)}$
and $\beta$ being the Brody parameter which describes the transition between ordered ($\beta = 0$) and chaotic behaviour ($\beta = 1$). 
For the value $\beta = 0$ the Brody distribution is equivalent to the Poisson distribution and for the $\beta = 1$, 
we have equivalence to the Wigner surmise distribution \cite{WignerSurmise}, describing the chaotic regime of the Gaussian Orthonormal Ensembles.

\section{Results}

The goodness of fit of the analytic Brody formula (\ref{BrodyFormula}) with the distribution of the calculated mean distances between occurrences of a word is determined by calculating the mean square error (mse) value in a binned least-squares fit procedure. 
The number of bins in the histogram used to fit the Brody's distribution is 300 and the maximum unfolded distance is 5.  

After fitting the Brody distribution to the measured distance distributions (X), the mse fitting error and the beta distribution parameter are obtained. 
We define a set of stop words by selecting those words for which the fitting error mse is smaller than a specified threshold value (mse hreshold). 

As mentioned in the introduction, it is very difficult to determine what word is non-informative for the analyzed class of documents.
Usually, the determination of non-informative words and their removal from the analyzed text belongs to the preliminary part of text processing before its actual analysis (for example, classification, sentiment determination, summarization). Only after calculating the final analysis metrics we are able to determine whether by removal of the given set of stop words we were able to increase the accuracy of the analysis or not. \\
Therefore, before conducting the main analysis, different sets of stop words corresponding to different mse matching error thresholds should be defined. 
Carrying out the main analysis of the texts and evaluating the main metric for various stop word variants will allow to determine the best threshold value of the mse fitting error.

For all languages analyzed in this work, the limiting value of the matching error mse was determined using $10\%$ percentile value of all mse obtained during calculations of fits of the Brody distribution. The mse thresholds, the number of obtained stop words and first 100 words for the selected languages are summarized the table \ref{table1}.

\begin{table}[h!]
\centering
\begin{tabularx}{\textwidth} { 
  | p{2cm} 
  | p{2cm} 
  | p{2cm}  
  | >{\raggedright\arraybackslash}X |} 
 \hline
 language & mse error threshold&  number of stop words & first 100 stop words \\ 
 \hline
 \hline
 English & 0.0094 & 975 & {\footnotesize in,an,th,re,on,he,the,al,or,at,is,of,to,st,it,me,and,as,co,de,la,us,
so,no,et,be,for,ad,her,by,iv,we,all,per,art,his,man,men,with,are,
use,from,reference,if,one,references,red,also,was,up,other,end,
out,own,age,that,can,act,which,not,our,led,see,form,this,ov,er,
son,but,here,now,any,has,low,part,ever,king,its,port,late,time,
war,rate,met,she,two,car,first,may,old,used,side,some,have,
nation,who,been,new,when,into,after} \\ 
 \hline
 German & 0.0145 & 907 & {\footnotesize  er,de,in,st,der,an,es,al,le,ein,ich,und,us,et,la,am,ist,von,den,im,
die,eine,da,des,um,ab,so,zu,to,aus,sie,mit,auf,ende,nach,auch,
als,dem,du,für,das,bei,ob,sei,sich,man,vor,unter,einer,wie,erst,
wo,über,art,ort,ins,iv,links,je,chr,weise,reich,dies,durch,zur,war,
bis,einen,jahr,teil,oder,zum,diese,hin,weit,nie,pro,alle,nun,zeit,
einem,sein,wurde,wird,alte,ca,land,lag,nicht,erste,werden,hat,
acht,andere,dar,kann,sind,neu,seit,zwei} \\
 \hline
 Polish & 0.03 & 888 & {\footnotesize  na,te,po,ze,go,ta,od,on,za,ego,nie,to,do,je,ma,la,we,tego,ok,tu,
si{\k{e}},przy,j{\k{a}},im,pod,\.{z}e,czy,mu,ii,ona,dni,co,ich,ale,raz,przez,oni,
one,jest,ul,jak,zm,sta{\l},niej,t{\k{e}},tak,stan,by{\l},rok,oraz,a\.{z},te\.{z},czas,
lat,{\'{s}}w,lub,zosta{\l},praw,który,nad,jako,kt\'{o}re,tym,przed,roku,s{\k{a}},
sam,dla,jego,ten,nowa,trzy,vi,tak\.{z}e,jedna,mi{\k{e}}dzy,i\.{z},lata,gen,
san,rp,ang,tych,polski,tej,bez,miast,dom,tam,nim,r\'{o}wnie\.{z},
cz{\k{e}}\'{s}ci,jednak,rz{\k{a}}d,jej,plik,dnia,stanowi,latach,pierwszy}\\
 \hline
\end{tabularx}
\caption{Details of extracted sets of stop words. The mse error threshold corresponds to the $10\%$ percentile of the mes value distributions. In most cases, the words presented in German and Polish do not correspond to the English words. Due to lack of space, we have decided not to translate them.}
\label{table1}
\end{table}

To illustrate the fits of the Brody distribution to the found word distance distributions, Figures 1,2,3 show for each language a sample of 4 distributions identified as stop words.  

\begin{figure}[h!]
  \centering
  \begin{subfigure}[b]{0.49\linewidth}
    \includegraphics[width=\linewidth]{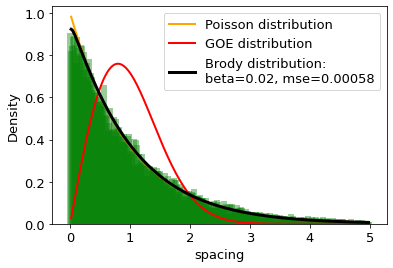}
    \caption{word:  {\it in}}
  \end{subfigure}
  \begin{subfigure}[b]{0.49\linewidth}
    \includegraphics[width=\linewidth]{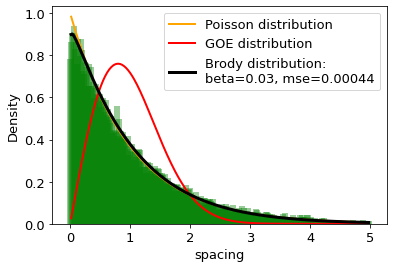}
    \caption{word:  {\it and} }
  \end{subfigure}
  \begin{subfigure}[b]{0.49\linewidth}
    \includegraphics[width=\linewidth]{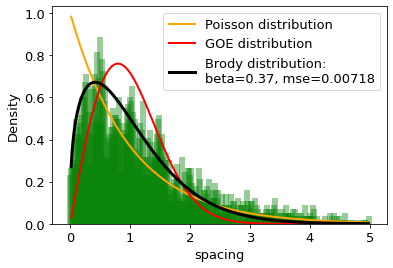}
    \caption{word:  {\it itself}}
  \end{subfigure}
  \begin{subfigure}[b]{0.49\linewidth}
    \includegraphics[width=\linewidth]{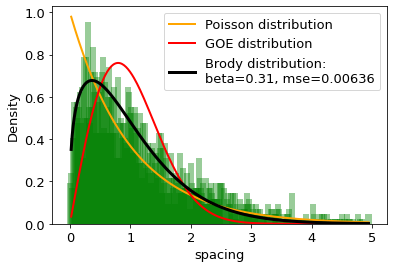}
    \caption{word:  {\it whom} }
  \end{subfigure}
  \caption{Calculated distributions of the average distance between words in word units for selected words in English. 
For each word 3 different distributions are shown. Poisson distributions (yellow line) and GOE (Wigner surmise, red line) illustrate extreme cases. The calculated fit of the Brody distribution is indicated by the black line. Number of bins for each histogram is 300. The best matched $\beta$ parameter ("beta") and error mse are noted in the legend of the image 
describing the fit of Brody formula. }
\label{english_set}
\end{figure}

\begin{figure}[h!]
  \centering
  \begin{subfigure}[b]{0.49\linewidth}
    \includegraphics[width=\linewidth]{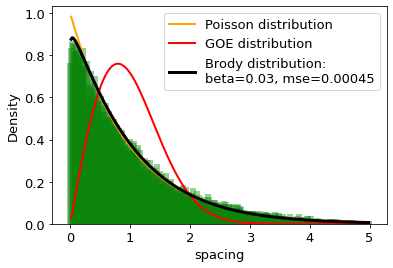}
    \caption{word:  {\it und (and)}}
  \end{subfigure}
  \begin{subfigure}[b]{0.49\linewidth}
    \includegraphics[width=\linewidth]{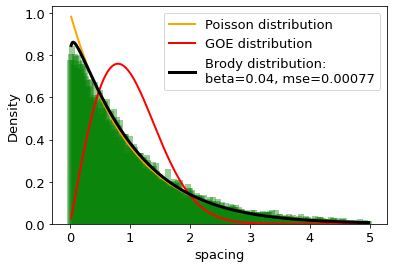}
    \caption{word:  {\it in (in)} }
  \end{subfigure}
  \begin{subfigure}[b]{0.49\linewidth}
    \includegraphics[width=\linewidth]{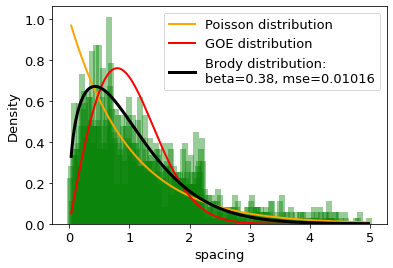}
    \caption{word:  {\it folgenden (following)}}
  \end{subfigure}
  \begin{subfigure}[b]{0.49\linewidth}
    \includegraphics[width=\linewidth]{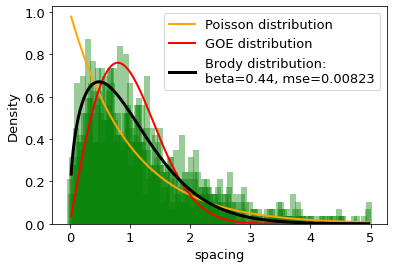}
    \caption{word:  {\it stellt (provides)} }
  \end{subfigure}
  \caption{Calculated distributions of the average distance between words in word units for selected words in German. 
Figure description as in Fig \ref{english_set} . The English translation is placed in brackets next to the German word. }
\label{german_set}
\end{figure}

\begin{figure}[h!]
  \centering
  \begin{subfigure}[b]{0.49\linewidth}
    \includegraphics[width=\linewidth]{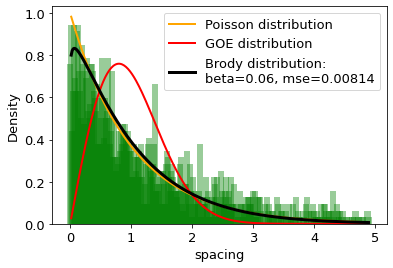}
    \caption{word:  {\it \'{s}w (St., Saint )}}
  \end{subfigure}
  \begin{subfigure}[b]{0.49\linewidth}
    \includegraphics[width=\linewidth]{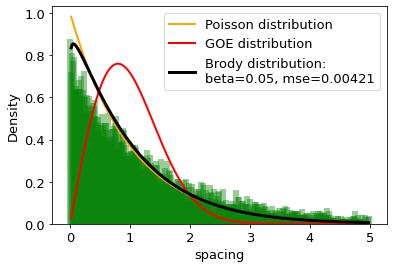}
    \caption{word:  {\it do (to)} }
  \end{subfigure}
  \begin{subfigure}[b]{0.49\linewidth}
    \includegraphics[width=\linewidth]{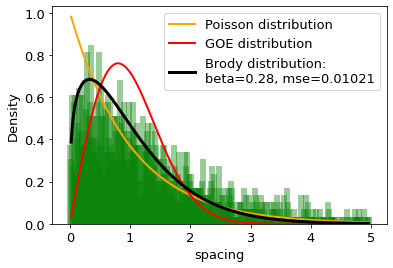}
    \caption{word:  {\it wiele (much)}}
  \end{subfigure}
  \begin{subfigure}[b]{0.49\linewidth}
    \includegraphics[width=\linewidth]{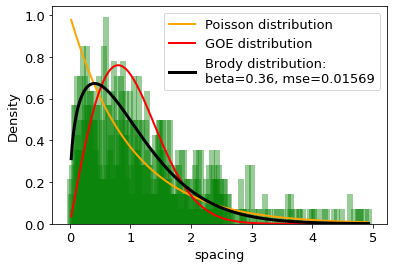}
    \caption{word:  {\it gdy\.{z} (because)} }
  \end{subfigure}
  \caption{Calculated distributions of the average distance between words in word units for selected words in Polish. 
Figure description as in Fig \ref{english_set} . The English translation is placed in brackets next to the Polish word. }
\label{polish_set}
\end{figure}

For each language, the first 2 subfigures ((a),(b))
show the most Poisson-like distributions. The remaining two ((c),(d)) show the state in which regular and chaotic regimes coexist. 
The greater the value of the beta parameter, the greater the share of the chaotic component. 

\section{Concluding Remarks}

The proposed method of customized determination of stop words for a given set of documents, is based on results of the Random Matrix Theory, in particular 
on the fitting of the Brody distribution to the calculated distribution of average distances between a given word in the set of texts.
Distance averaging is performed independently for each text.
Goodness of fit, in our case it is the mean square error, is a parameter by means of which it is defined which word belongs to the stop words set. If the match error is less than the specified value (mse threshold), then the word is determined as stop word. 

The usage of Wikipedia articles ensures that texts are edited by different authors. Therefore, the style of the articles describing the various subjects is different. This translates into different characteristics of the occurrence of individual words in sentences and their relationships with other words, different length of sentences, etc.
Hence, we conclude that the obtained mean distance distributions for the most common words illustrate well-generalized results, which confirms the agnostic nature of the proposed method. 

A very important feature of our method for identifying uninformative words is that we do not need to perform a lemmatization step. This part of data preprocessing is very difficult for most languages. Therefore, eliminating this part in data preparation is a great simplification of the whole analysis task. 

\clearpage
\section*{Acknowledgement}
I am grateful to Antonia {\L}obodzi\'{n}ska for critical comments.


\begin{thebibliography}{99}
\bibitem{Wikiarticles} https://dumps.wikimedia.org/plwiki/latest, https://dumps.wikimedia.org/enwiki/latest, https://dumps.wikimedia.org/dewiki/latest
\bibitem{PhysicalRMT} E. Brezin, V. Kazakov, D. Serban, P. Wiegmann, A. Zabrodin, (2004), Applications of Random Matrices in Physics 
\bibitem{BiologyRMY} Kikkawa, A. Random Matrix Analysis for Gene Interaction Networks in Cancer Cells. Sci Rep 8, 10607 (2018). https://doi.org/10.1038/s41598-018-28954-1
\bibitem{FinanceRMT} Complex market dynamics in the light of random matrix theory
Hirdesh K. Pharasi, Kiran Sharma, Anirban Chakraborti, Thomas H. Seligman,arXiv:1809.07100 q-fin.ST
\bibitem{Unfolding} Improved unfolding by detrending of statistical fluctuations in quantum spectra
Irving O. Morales, E. Landa, P. Str\ {a}nsk\ {y}, and A. Frank, Phys. Rev. E 84, 016203
\bibitem{EMD} https://pypi.org/project/EMD-signal
\bibitem{empyricalRMT} https://pypi.org/project/empyricalRMT
\bibitem{Berry} M. V. Berry and M. Tabor, Level clustering in the regular spectrum,Proc. Roy. Soc.A356(1977) 375–394.
\bibitem{WignerSurmise} M. L. Mehta, Matrices and the statistical theory of Energy levels (Academic Press, New York) 1967
\bibitem{Brody} T.A.Brody, Lett. Nuovo Cimento, 7 (1973), 482
\end{thebibliography}
\end{document}